# Domain Knowledge Injection in Bayesian Search for New Materials

Zikai Xie[a;*], Xenophon Evangelopoulos[a,b], Joseph C. R. Thacker[c] and Andrew I. Cooper[a,b]

[a]Department of Chemistry, University of Liverpool, Liverpool, UK
[b]Leverhulme Research Centre for Functional Materials Design
[c]Daresbury Laboratory STFC UKRI, Daresbury, UK
ORCiD ID:

**Abstract.** In this paper we propose DKIBO, a Bayesian optimization (BO) algorithm that accommodates domain knowledge to tune exploration in the search space. Bayesian optimization has recently emerged as a sample-efficient optimizer for many intractable scientific problems. While various existing BO frameworks allow the input of prior beliefs to accelerate the search by narrowing down the space, incorporating such knowledge is not always straightforward and can often introduce bias and lead to poor performance. Here we propose a simple approach to incorporate structural knowledge in the acquisition function by utilizing an additional deterministic surrogate model to enrich the approximation power of the Gaussian process. This is suitably chosen according to structural information of the problem at hand and acts a corrective term towards a better-informed sampling. We empirically demonstrate the practical utility of the proposed method by successfully injecting domain knowledge in a materials design task. We further validate our method's performance on different experimental settings and ablation analyses.

## 1 Introduction

Observing physical phenomena often requires a careful design of experiments [5] to gain useful insights and discover new knowledge. Such a design can be impractical due to the combinatorially large and convoluted space of choices one has to consider. Bayesian optimization (BO) [16, 31] has emerged as a sample-efficient agent to optimize over large spaces through iteratively querying potentially unknown, expensive to evaluate objectives (i.e., black-box) that often involve noisy measurements. BO has been successful across various disciplines including tuning machine learning [3, 22], robotics [21], online learning [34] reinforcement learning [26], selection of chemical compounds [12] and in the design of new materials [6].

Formally BO seeks to find the global optimum in the following problem

$$\mathbf{x}^\star = \arg\max_{\mathbf{x} \in \mathcal{X}} f(\mathbf{x}), \qquad (1)$$

where $f : \mathcal{X} \to \mathbb{R}$ is a function on the compact subset $\mathcal{X} \subseteq \mathbb{R}^d$. Usually no formulation or information over $f(\cdot)$ is given, but we rather capture our beliefs about its behavior through a prior distribution which is progressively updated as new data is acquired. BO acts iteratively with the first step involving the selection of a *surrogate model* that approximates the correlation of the observed measurements with the parameter space. The next step requires the design of mechanisms to determine new points to query the objective in each iteration, i.e., designing an *acquisition function*. This normally involves calculating the expected informativeness of learning $f(\mathbf{x})$ for every point and in practice acts as a trade-off between exploration versus exploitation in the sampling behavior of the optimizer. As new measurements are collected, the surrogate model is updated accordingly and this two-step procedure is repeated until convergence. Gaussian processes (GP) [29] are typically used as surrogate models in most BO instantiations mainly due to their efficiency and simplicity, i.e., having an amenable analytical form of the posterior distribution. They form a collection of random variables where the joint distribution of any finite subset is still a Gaussian distribution. In practice, a zero-mean GP prior is often employed and some common choices for a covariance kernel are the square exponential kernel and Matérn kernel [29].

Most recently there has been an increasing interest over how to exploit external knowledge in BO towards accelerating convergence [28]. Many of these contributions primarily focus on employing external knowledge in a form of prior distribution added to the model towards guiding the optimization to more fruitful regions [36, 19, 15]. In most scientific tasks however, external knowledge cannot always be fairly translated in a form of prior distribution. More importantly, such strategies hinder the induction of prior bias in the problem and may ultimately shift the search into unexciting regions. Another series of works attempts to instead introduce domain-specific knowledge or rules in a form of constraints [11].

In practice, it may make more sense to introduce other types of external knowledge to the problem, such as structural information about the optimization space or about the relationship between the input parameters and the target objective, e.g., linear or non-linear, etc. This type of domain knowledge can be more amenable for physical science problems such as materials science for example where an optimal linear mixture of chemical compositions is sought [6]. In such domains especially, due to the inherent extreme imbalance of optimality conditions, most surrogate models resort to smoothing over the optimum or over-predicting near its location which can often result to a local-minima confinement [35, 32]. While other choices of surrogate models have been used in BO such as Random forests [14] and Bayesian neural networks [37], these have empirically shown to predominate exploration and lead to poor performance [7].

---

* Corresponding Author. Email: zikaix@liverpool.ac.uk

Instead, in this work we propose a simple, yet novel approach that can capture such domain-specific knowledge in BO by adaptively augmenting the acquisition function with a suitably chosen predictive model trained iteratively over previously sampled points in an online learning fashion [22]. Instead of using a predictive model as a surrogate model, we integrate its prediction power in the acquisition function as a corrective term, allowing the benefits of both GP and predictive model to be utilised. The proposed approach allows for more flexibility in the choice of acquisition functions and appears to improve upon standard BO on a materials design simulation task when suitable prognostic models are chosen. Extensive empirical results and ablation analyses demonstrate that our method maintains a competitive and robust performance overall.

The rest of the paper is organised as follows. Section 2 presents recent works related to external knowledge exploitation in BO, while Section 3 outlines the details of the proposed methodology. The performance and robustness of our algorithm is evaluated and discussed in Section 4 and Section 5 concludes our work and highlights some future directions.

## 2 Relation to existing methods

External knowledge injection in BO has recently become fashionable with a number of interrelated approaches having been proposed over the last few years. A common issue in BO in general is the so-called "cold start" problem, i.e., the initial, usually randomly chosen points, fail to adequately capture the landscape of the optimizing objective. Most recently, a new branch of research explored the challenging task of utilizing knowledge from prior BO "campaigns", e.g., meta-learning [23], to help warm-start the optimization as well as injecting external information [17]. Transfer learning has been successfully applied to chemical reaction optimization [13] to bias the search space by weighting the acquisition function of the current campaign with past predictions.

A major line of works propose BO frameworks that incorporate external knowledge in a form of prior beliefs from a fixed set of distributions [3]. A number of variants have been proposed that can accommodate generative models coupled with user defined priors into pseudo-posteriors [36] where the prior distribution is modeled using the "positive observations" for more efficient sampling. Other works incorporate prior user beliefs with observed data and compute the posterior distribution through repeated Thompson sampling [19]. New sampling points are approximated using a linear combination of posterior samplings. In [28] prior beliefs are used to highlight high-probable regions in terms of optimality through the probability integral transform method. Finally, the work of [15] adaptively integrates prior beliefs in the acquisition function as a decayed multiplicative term towards improved sampling, maintaining at the same time standard acquisition function convergence guarantees. The use of external knowledge as a GP prior provides a means to correct GP predictions; however the customized mean function tends to dominate as the optimization progresses in practice [30] and has shown to negatively affect the performance overall [7].

While the accommodation of expert beliefs either as a surrogate or in a form of acquisition function has been well studied recently, structural aspects of the optimization have been at the forefront of knowledge injected BO. In particular such methods employ structural priors, other than the GP kernel to model how the objective function is expected to behave [14]. Such priors can either model the monotonicity of the objective [20] or its non-stationarity [34]. Another series of works attempts to alleviate the issue of overexploring the boundaries of the search space using multi-task Gaussian process [38]. The work of [25] proposes a cylindrical kernel that expands the center of the search space and shrinks the edges, while [33] propose adding derivative signs to the edges of the search space to steer BO towards the center. Nevertheless, most of the aforementioned methods address specific only structural aspects tailored to the problem at hand that are not directly generalizable. In this work we instead propose a generalized strategy for structural knowledge injection where a suitable predictive model is used to augment the acquisition function and enrich the search with structural properties of the problem at hand towards more effective exploitation in the optimization.

## 3 Enriching acquisition functions with domain knowledge

In this section we propose a straightforward approach to inject general type of external knowledge in BO by augmenting the acquisition function through a tunable predictive model acting as an assistive surrogate to enrich the approximation power of the Gaussian process model. Let $\mathcal{D} = \{(\mathbf{x}_i, y_i)\}_{i=1}^n$ be an observation dataset and $\alpha(\mathbf{x}, \mathcal{D})$ be an acquisition function, i.e., a criterion used to obtain new candidate samples to evaluate across the various iterations of BO. A commonly used acquisition function is upper confidence bound (UCB) defined as

$$\mathbf{x}^\star = \arg\max_{\mathbf{x} \in \mathcal{X}} \mu(\mathbf{x}) + \kappa \sigma(\mathbf{x}), \qquad (2)$$

where $\mu(\mathbf{x})$ is the posterior prediction, $\sigma(\mathbf{x})$ is the uncertainty and the total objective represents a trade-off between exploration versus exploitation in the search. We now enrich Eq. (2) with the output of a predictive model $\xi(\mathbf{x}, \mathcal{D})$ such as a random forest or any other tree-based model [14] at $\mathbf{x}$ as

$$\mathbf{x}^\star = \arg\max_{\mathbf{x} \in \mathcal{X}} \alpha(\mathbf{x}, \mathcal{D}) + \gamma \xi(\mathbf{x}, \mathcal{D}). \qquad (3)$$

That is, the assistive predictive model is being trained sequentially on new sample points as these arrive independently from the BO procedure in a self-supervised regime. The addition of the prediction output itself works as a correction to the sampling space without affecting the parameters of the original Gaussian process. This can be a suitably chosen deterministic model capturing various structural information of the problem at hand towards a better-informed sampling. Figure 1 further exemplifies the effects of augmenting the acquisition in Eq. (3) by showing the different approximations of the Ackley function by a GP and random forest regressor, as well as their combined approximation power. Evidently, the latter panel demonstrates a better approximation power by the new acquisition and thus a richer sampling strategy.

While this iterative training process can incur a running-time offset, especially as the number of samples increases, this practically becomes negligible in real-world tasks where measuring physical phenomena is a time-consuming process. We further discuss and exemplify this later in the experiments section where we demonstrate that the proposed approach inhibits no delay in a materials design optimization problem.

Depending on the choice of acquisition function, a weighting parameter $\gamma$ needs to be chosen to ensure the predictive model term is up to scale, without however affecting the prediction output of the Gaussian process. For UCB, we select $\gamma = 1$ as the predictive model output naturally acts as a correction to the posterior mean of the GP, being at a similar scale. For EI and POI, which reflect the possible improvement on the best score, a suitably designed scaling factor is used to normalise the predictive contribution and bring the two terms in the acquisition function to scale as follows

$$\gamma = \frac{\sum \alpha(\mathbf{x}_{init}, \mathcal{D})}{\sum \xi(\mathbf{x}_{init}, \mathcal{D})}, \tag{4}$$

where $\mathbf{x}_{init}$ stands for the initial set of points used to seed the optimizer. This can be selected either randomly or through a careful design of experiments.

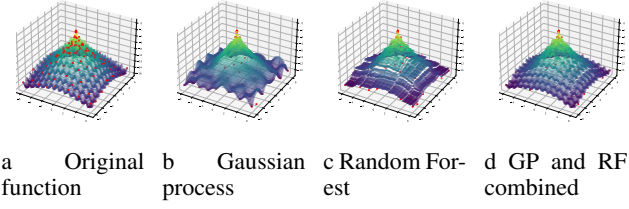

| a Original function | b Gaussian process | c Random Forest | d GP and RF combined |

**Figure 1**: Approximation of the Ackley function by two different surrogate models and their combination. Red points represent datapoints upon which the models are trained.

In practice, the utility of the predictive model term tends to be negligible during the initial steps of the optimization due to lack of training data. Therefore, we employ a monotonically increasing weight for the correction term that deemphasizes its contribution in the initial stages, allowing a more GP-dominated sampling to take effect, while boosting its contribution further down the optimization process as soon as enough training data is made available. This can be achieved by introducing the following quadratically increasing weight function normalized to be in [0,1]

$$h(i) = min\left(1, \frac{4i^2}{i_{max}^2}\right), \tag{5}$$

where $i$ denotes the optimization iteration number and $i_{max}$ is the predefined total number of iterations. We therefore adjust the proposed acquisition function as follows

$$\widetilde{\alpha}(\mathbf{x}, \mathcal{D}, i) = \alpha(\mathbf{x}, \mathcal{D}) + \gamma h(i) \xi(\mathbf{x}, \mathcal{D}). \tag{6}$$

Furthermore, we have empirically observed that depending on the problem at hand, the predictive model in the acquisition function can dominate the search over the GP model and get stuck into local optima affecting overall performance. To address this issue we further propose an early stopping strategy so that the predictive model can be dropped when such phenomena are observed, allowing the optimizer to carry on only using standard GP prediction in a warm-start fashion. We particularly monitor such cases by examining the closeness of the different sampling points between consecutive BO iterations. The following condition describes our proposed early stopping criterion

$$\frac{\|\mathbf{x}_i - \mathbf{x}_{i+1}\|}{\left\|\mathbf{x}_{i+1} - \frac{1}{i}\sum_{k=1}^{i}\mathbf{x}_k\right\|} < \epsilon. \tag{7}$$

Algorithm 1 summarizes the main steps of our proposed algorithm, termed throughout the rest of the paper as DKIBO.

## 4 Experimental comparisons

We present a set of experiments to demonstrate the practical utility of the proposed method on a wide range of problems, including the physical-world task of searching for new materials. We empirically compare performances across various methods and demonstrate the robustness of the proposed method. Section 4.1 details the various experimental settings along with the selected comparison methods used for benchmarking. Sections 4.2, 4.3, 4.4 and 4.5 present results for four experimental settings, namely an analytical function optimization task [7], a hyperparameter optimization task [1], a robotic swimming simulation task [39] and a materials mixture design problem [6]. Finally, Section 4.6 details some ablation studies that further highlight the robustness of the proposed methodology.

**Algorithm 1** Domain Knowledge Injected Bayesian Optimization
**Require:** Acquisition function $\widetilde{\alpha}$, maximum iteration number $i_{max}$, prediction model $\xi(\mathbf{x}, \mathcal{D})$, $\epsilon = 0.05$.
1: Initialize observation dataset by random sampling $\mathcal{D} \leftarrow \{(\mathbf{x}_{init}, y_{init})\}$;
2: **if** $\alpha(\cdot)$ is UCB **then**
3:    $\gamma \leftarrow 1$;
4: **else**
5:    $\gamma \leftarrow \frac{\sum \alpha(\mathbf{x}_{init}, \mathcal{D})}{\sum \xi(\mathbf{x}_{init}, \mathcal{D})}$;
6: **end if**
7: $i \leftarrow 0$;
8: **while** $i < i_{max}$ **do**
9:    Fit the predictive model $\xi(\mathbf{x}, \mathcal{D})$ with $\mathcal{D}$;
10:    Fit the Gaussian process $GP \sim \mathcal{N}(0, \mathbf{K}_{\text{Matérn}})$ with $\mathcal{D}$;
11:    Maximize $\widetilde{\alpha}(\mathbf{x}, i)$ and get new observation point $\mathbf{x}^\star$;
12:    Probe $\mathbf{x}^\star$ to get $y^\star$ and $\mathcal{D} \leftarrow \mathcal{D} + \{(\mathbf{x}^\star, y^\star)\}$;
13:    **if** $\frac{\|\mathbf{x}_i - \mathbf{x}_{i+1}\|}{\left\|\mathbf{x}_{i+1} - \frac{1}{i}\sum_{k=1}^{i}\mathbf{x}_k\right\|} < \epsilon$ **then**
14:      $\gamma \leftarrow 0$;
15:    **end if**
16:    $i \leftarrow i + 1$;
17: **end while**

### 4.1 Experimental setup

We empirically test the performance of the proposed algorithm across the following tasks:

- **Synthetic functions** Synthetic benchmark functions (as used in [7]) of varying dimensionality with multiple local optima used to test precision, convergence speed and robustness of BO methods. Simple regret and cumulative mean regret are used to measure the performance of the optimization result and convergence speed, respectively. The maximum number of iterations is set to 100 and the median value of 50 repeated trials is reported.
- **BBO challenge** Black-box optimization challenge (NeuIPS 2020) [1] is a hyperparameter tuning challenge over a large number of machine learning models such as SVM, random forest, etc. The score is computed by the bayesmark package using the minimum value achieved by an algorithm normalized by the expected minimum and maximum objective function values $f$, following the equation: $s_i = 100 * \frac{f_i - f}{f_i - f}$.
- **Robotic swimming task** Robotic swimming simulation environment MuJoCo [39] that is used for testing reinforcement learning tasks, where the goal is to swim as fast as possible. Here we adapt the task to a black-box optimization one by linearly mapping the observation space matrix to the action space as a 16-dimensional optimization problem where the target is the average reward. The maximum number of iterations is set to 100 and the median value of 50 repeated trials is reported.

- **Photocatalytic hydrogen production** We replicate the materials design problem addressed in [6] to maximize photocatalytic hydrogen evolution rate (HER) out of mixture of different materials. Here, we employ a rather cost-efficient approach where new HER measurements are interpolated through a multi-layered auto-ML ensemble model of neural networks, namely Autogluon [8]. Section 4.5 provides a detailed description of this task. The maximum number of iterations is set to 100 and the median value of 50 repeated trials is reported.

We experiment with the following optimizers:

- **Standard Bayesian optimization (SBO)** Standard BO algorithm [24] employing a scikit-learn [27] GP with Matern kernel ($\nu = 2.5$) optimized with L-BFGS-B algorithm [18] by the SciPy package [40], warm-started with 5 initial points.
- **Sequential model-based algorithm configuration (SMAC)** A surrogate-based optimization algorithm [14] using random forest as its surrogate model. Random sampling or variance from trees is used to acquire new points. The SkOpt package is used here [10].
- **Tree-structured Parzen estimator (TPE)** TPE [3] uses a truncated Gaussian mixture with two densities to model $p(\mathbf{x}|y)$ and calculate $p(y|\mathbf{x})$ according to the Bayesian rule. We use the implementation provide by [4].
- **Opentuner (OT)** Opentuner[2] is based on the idea of stacking; several search techniques are included as ensembles to allocate probe points, while AUC bandit algorithm controls the weight and number of different search techniques which allocate. The search techniques include differential evolution, many variants of Nelder-Mead search and Torczon hillclimbers, a number of evolutionary mutation techniques, pattern search, particle swarm optimization and random search.
- **Random search (RS)** Random search under uniform distribution over the search space.

Throughout our analyses, we use preset hyperparameters for each of the compared methods. For all GP-based models, including the proposed, a Matérn kernel ($\nu = 2.5$) with added white noise, optimized with L-BFGS-B algorithm [18] and warm-started with 5 initial points is used. Random seed for each trial is set to the trial number. The default acquisition function is UCB with $\kappa$ set to 2.6. The benchmark results and code can he found in https://github.com/XieZikai/DKIBO.

### 4.2 Synthetic functions

In this section we experiment with a set of synthetic set functions [7] of varying landscapes and dimensionalities to test the performance of the proposed method on different objectives with potentially large number of local optima. Table 1 presents the simple regret which measures the optimality performance of the different algorithms while Table 2 shows the cumulative mean regret which evaluates the convergence speed. The median value across all repeats along with standard deviation is reported in both tables. Here we enrich the DKIBO acquisition functions with a random forest regression model using 20 estimators with maximum depth of 5 tree splits and an early stopping hyperparameter set to 0.05.

The proposed method appears to perform best in terms of simple regret while maintains a competitive performance in terms of cumulative mean regret across of a set of diverse synthetic functions, as shown in Table 2. Compared to standard BO, DKIBO offers a consistent improvement on both measures.

### 4.3 BBO challenge

Table 3 reports the BBO score for each method. For standard BO and the proposed method we report performances over three different acquisition functions. In this experiment we enrich the DKIBO acquisition functions with a random forest regression model using 20 estimators with maximum depth of 5 tree splits and an early stopping hyperparameter set to 0.05. Evidently, the enriched acquisition function of DKIBO improves upon the standard ones which further highlights the generalizability of the proposed methodology. Overall, SMAC outperforms but DKIBO (using EI) appears to achieve the second best score, demonstrating a very competitive performance.

### 4.4 Robotic swimming task

We now test our algorithm on the MuJoCo [39] swimmer environment, which is a physical simulation of a 3D robotic swimmer with three segments and two articulation joints (rotors) to connect two of the segments to form a linear chain. The goal is to move forward as fast as possible by applying torque on the rotors using fluids friction. The environment contains a 2-dimensional action space for the rotors and 8-dimensional observation space. Since the environment is specifically designed for reinforcement learning tasks rather than black-box optimization, we wrap the problem using a (2, 8) matrix as a linear mapping weight from observation space to action space so that the problem transforms to a 16-dimensional black-box optimization problem where the target is to maximize the average reward.

Figure 2 demonstrates DKIBO's competitive performance against other compared methods. Here we again use a random forest regression as predictive model using 20 estimators with maximum depth of 5 tree splits and an early stopping hyperparameter set to 0.05. It is evident that DKIBO maintains very competitive performance in terms of convergence and simple regret, indicating that the optimization process generally benefits from the augmented knowledge. Notably, a plateauing effect on simple regret can be observed in the last few stages of the optimization in the right-most panel of Figure 2 which could be owed to a possible local optima confinement. Table 4 further details DKIBO's performance in terms of simple regret and cumulative mean regret.

### 4.5 Photocatalytic hydrogen production optimization

In this section we take on a materials design problem [6] where the goal is to find an optimal composition of materials that maximizes hydrogen production through photocatalysis [41]. In this experiment 10 different materials on various concentrations were used increasing the search space (full simplex) to 98,423,325 possible combinations. Due to the increasing demand for tractable optimizers for costly real-world tasks, BO has emerged as a competitive method in the physical sciences community. Here, we recast the problem in a simulated fashion and take advantage of the underlying linear relationship between the input and output parameters (*structural knowledge*). We specifically enrich our acquisition function with a linear regression model trained by a least-squares error and simulate the HER measurements by fitting existing lab measurements reported in [6] using Autogluon [8], a strong ensemble model that can approximate new measurements at new sampled points. While the simulated measurements are only approximate, our goal here is to successfully utilise underlying structural knowledge as a proof-of-concept.

Figure 3 demonstrates the dominance of BO-based algorithms over other baseline algorithms in terms of HER and simple regret.

| | DKIBO | SBO | TPE | RS | OT | SMAC |
|---|---|---|---|---|---|---|
| Colville$_{d=5}$ | **31.61±20.87** | 39.52±22.55 | 383.28±353.20 | 813.56±652.85 | 157.90± 269.58 | 825.70±636.69 |
| Michalewicz$_{d=10}$ | 3.57 ± 0.61 | 3.94 ± 0.59 | 4.16 ± 0.44 | 4.46 ± 0.41 | **2.41±0.87** | 3.74 ± 0.50 |
| Ackley$_{d=2}$ | **0.13±0.28** | 0.31±0.50 | 5.76±1.79 | 8.02±2.51 | 2.71±2.23 | 2.65 ± 5.47 |
| Branin$_{d=2}$ | 6.75×10$^{-5}$±1.49×10$^{-4}$ | **6.67×10$^{-5}$±1.17×10$^{-4}$** | 0.17±0.24 | 0.26±0.34 | 0.018±0.28 | 0.073±0.65 |
| Eggholder$_{d=2}$ | **1.12±4.30** | 1.92±3.55 | 13.79±29.36 | 83.76±54.51 | 8.77±50.68 | 35.86±236.81 |
| Goldstein price$_{d=2}$ | **2.86 ± 4.80** | 3.04±4.30 | 2.92±8.47 | 12.12±18.63 | 0.63±26.77 | 33.85±74.01 |
| Hartmann$_{d=6}$ | 0.012±0.24 | **1.67×10$^{-3}$±0.061** | 0.77±0.28 | 1.33±0.40 | 0.14±0.13 | 0.33±0.41 |
| Rosenbrock$_{d=2}$ | **0.10±0.19** | 0.12 ± 0.18 | 2.38±4.25 | 3.75±10.63 | 0.95±4.09 | 4.82±9.72 |
| Six hump camel$_{d=2}$ | 3.14×10$^{-4}$±4.42×10$^{-3}$ | 1.26×10$^{-3}$±0.013 | 0.049±0.074 | 0.14±0.16 | 6.06×10$^{-3}$±0.094 | 8.24×10$^{-3}$±0.23 |
| StyblinskiTang$_{d=2}$ | **7.14×10$^{-4}$± 9.04×10$^{-4}$** | 7.50×10$^{-4}$± 1.13×10$^{-3}$ | 0.99±4.34 | 2.76±4.30 | 0.21±4.49 | 1.49±9.4 |

Table 1: Simple regret performance on synthetic functions with various dimensionalities $d$. Best performances appear boldfaced.

| | DKIBO | SBO | TPE | RS | OT | SMAC |
|---|---|---|---|---|---|---|
| Colville$_{d=5}$ | 2851.95±2753.50 | 3003.76±2964.77 | 2908.38±2447.89 | 3134.14± 2224.16 | **2466.44±2227.98** | 3923.27±3044.18 |
| Michalewicz$_{d=10}$ | 4.47±1.09 | 4.60±0.94 | 4.65±0.71 | 4.79±0.54 | **3.57 ±1.57** | 4.570.96 |
| Ackley$_{d=2}$ | **2.96 ± 2.74** | 3.06±2.67 | 9.27±3.91 | 11.38±3.79 | 6.37±4.20 | 8.01±6.27 |
| Branin$_{d=2}$ | 1.42±1.42 | **1.33±1.33** | 1.69±1.49 | 2.13±1.81 | 1.92±1.82 | 2.00±1.77 |
| Eggholder$_{d=2}$ | 73.57±71.21 | **70.75±68.01** | 77.25±60.08 | 180.71±106.84 | 118.94±101.52 | 177.84±211.39 |
| Goldstein price$_{d=2}$ | 312.02±307.81 | 263.4±267.2 | 262.7±275.9 | **200.97±183.86** | 235.69±225.09 | 248.62±207.54 |
| Hartmann$_{d=6}$ | **0.70±0.63** | 0.71±0.68 | 1.41±0.69 | 1.73±0.60 | 0.86±0.70 | 0.95±0.62 |
| Rosenbrock$_{d=2}$ | **257.91 ± 257.75** | 285.44 ± 285.34 | 358.14±354.28 | 745.21±737.69 | 488.60±486.10 | 316.92±308.64 |
| Six hump camel$_{d=2}$ | 0.55±0.55 | **0.55±0.54** | 0.59±0.52 | 0.80±0.63 | 0.62±0.60 | 0.63±0.57 |
| StyblinskiTang$_{d=2}$ | **4.78±4.78** | 5.83±5.83 | 7.96±6.65 | 10.11±7.05 | 6.54±6.32 | 12.01±9.36 |

Table 2: Cumulative mean regret performance on synthetic functions with various dimensionalities $d$. Best performances appear boldfaced.

| | SMAC | OT | TPE | RS | SBO | | | DKIBO | | |
|---|---|---|---|---|---|---|---|---|---|---|
| | | | | | UCB | EI | POI | UCB | EI | POI |
| BBO score | **94.16** | 86.90 | 92.26 | 83.09 | 89.87 | 91.83 | 87.43 | 92.86 | 93.10 | 87.99 |

Table 3: BBO scores for different optimization algorithms. Larger scores are better and best performance appears boldfaced.

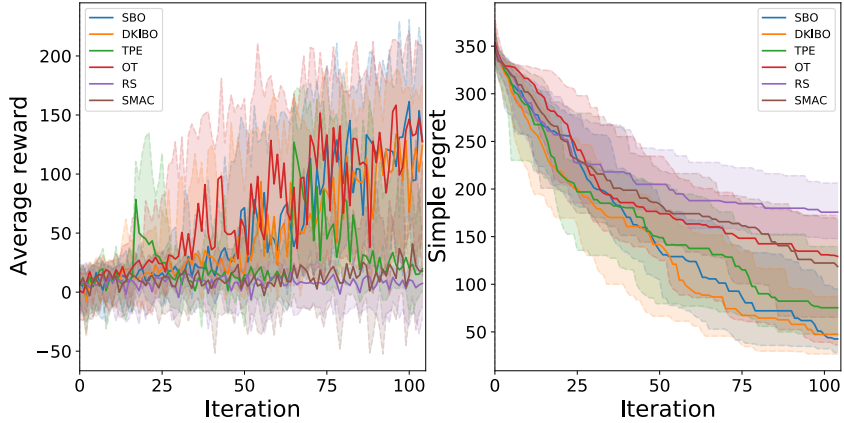

Figure 2: Average reward (left) and simple regret (right) performance comparison on the robotic swimming task. Solid lines show the median values while shaded areas represent the $[25, 75]$ percentile area.

| | CMR | Simple Regret |
|---|---|---|
| DKIBO | **142.81±94.45** | 47.29±60.53 |
| SBO | 155.35± 103.21 | **42.62±44.31** |
| TPE | 163.51±88.91 | 75.38±55.04 |
| OT | 191.33±107.92 | 129.26±79.13 |
| RS | 218.62±66.08 | 175.68±47.69 |
| SMAC | 194.85±95.70 | 118.64±59.58 |

Table 4: Simple regret and cumulative mean regret performances on the robotic swimming experiment. Best performance appears boldfaced.

| | CMR | Simple Regret |
|---|---|---|
| DKIBO | **5.57±3.57** | **1.95±2.14** |
| SBO | 7.76±5.12 | 2.72±2.62 |
| TPE | 10.94±0.50 | 10.54±0.28 |
| OT | 10.83±1.14 | 10.27±0.91 |
| RS | 11.22±0.41 | 10.89±0.24 |
| SMAC | 10.71±0.44 | 10.29±0.15 |

Table 5: Simple regret and cumulative mean regret performances on the simulated photocatalysis experiment. Best performance appears boldfaced.

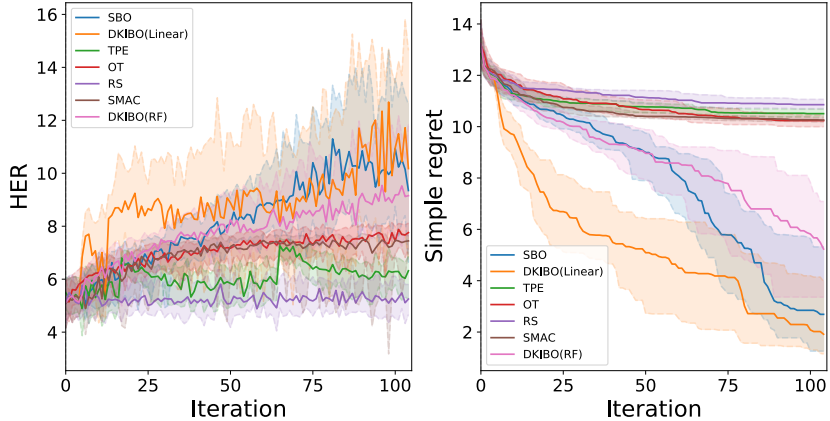

**Figure 3**: HER (left) and simple regret (right) performance comparison on the photocatalysis optimization task. Solid lines show the median values while shaded areas represent the [25, 75] percentile area.

Furthermore, the right-most panel highlights that the enrichment of DKIBO with linear regression significantly boosts performance in early stages and validates the utility of external knowledge injection in real-world problems, which is consistent with our observation in experiment 4.4. The left-most panel accordingly shows that the sample points suggested by the proposed approach are consistently better than standard BO (HER measurements are slightly different compared to the ones reported in [6] due to being interpolated by Autogluon; also, no batching is used here). At late stages however, the predictive model appears to induce fluctuations to the overall performance, which highlights the need for further improved adaptation strategies in future works. It is noteworthy that employing a random forest predictive model actually degrades DKIBO's performance, further highlighting the importance of injecting suitable knowledge for the task at hand. Table 5 further highlights the dominance of the proposed method in terms of simple regret and cumulative mean regret.

## 4.6 Ablation studies

In this section we perform a series of ablation studies and analyses to further investigate the performance and robustness of the proposed methodology.

### 4.6.1 Exploration of scaling effect in acquisition

We firstly explore the robustness of the proposed acquisition augmentation against scaling effects on the exploration aspects of the optimization. We specifically focus on the UCB acquisition function here mainly because its hyperparameter $\kappa$ directly controls the exploration scale. We empirically show that adding the predictive model term in the acquisition offers a consistent improvement on the model's overall performance regardless of the choice of the $\kappa$ hyperparameter in Eq. (2) in most cases. We vary $\kappa$ in $\{1.3, 2.6, 5.1\}$ and compare DKIBO with standard BO performance in Table 6 across a variety of settings.

To further demonstrate the flexibility of our proposed methodology we report performance using another predictive model, namely gradient boosted regression tree (GB) [9] based on 20 estimators and a maximum depth of 3. Due to space limitations we do not report performance using a linear predictive model here, as it empirically showed to only perform well for the photocatalysis experiment due to its underlying problem structure, and the best overall score was always achieved for $\kappa = 2.6$. Results demonstrate a consistent win of our methodology using a random forest predictive model (RF), with the gradient boosted (GB) showing a competitive performance across a variety of search spaces. This importantly highlights the ability of the added corrective term to learn to "rebalance" the exploration-exploitation trade-off under different parametrizations of the acquisition and emphasizes the potential to generalize to various predictive models depending on the problem at hand.

### 4.6.2 Comparison to a linear mean function GP

We further validate the utility of the added linear predictive model in the photocatalysis scenario by comparing against a simple GP surrogate with a linear mean function, highlighting major differences between the two approaches. A comparison in terms of both pure HER and simple regret exposes local optima confinement problems for the GP with a linear mean function which appears to dominate long-term predictions throughout the optimization process; a phenomenon also confirmed by [30]. We empirically confirm this here by employing an early stopping approach (ES) where the linear mean term is swapped with a zero-mean GP throughout the rest of the optimization based on the criterion of Eq. (7). HER and simple regret performance is shown in Figure 4 for the two linear mean variants and DKIBO. Evidently, a plataning effect is observed after iteration number 25 for the linear mean term BO in contrast to its respective ES variant, where a rapid performance boost follows after swapping to a zero-mean GP in that iteration. DKIBO on the other hand appears to perform best overall and maintain the correct balance between exploration versus exploitation in the acquisition term between UCB and the linear model.

### 4.6.3 Corrective term usage analysis

Finally, we report results on the average usage duration of DKIBO's adaptive corrective term to further investigate the utility of DKIBO's early stopping strategy. Figure 5 shows at which iteration the corrective term is being dropped by the early-stopping strategy proposed in Eq. (7). Evidently, in most cases the corrective term appears to be useful throughout the whole run, while in some cases appears to be dropped in early stages. Nevertheless, even an early stage use only appears to significantly boost performance acting as a warm-start for the optimization.

| | $\kappa = 5.1$ | | | $\kappa = 2.6$ | | | $\kappa = 1.3$ | | |
|---|---|---|---|---|---|---|---|---|---|
| | SBO | DKIBO$_{RF}$ | DKIBO$_{GB}$ | SBO | DKIBO$_{RF}$ | DKIBO$_{GB}$ | SBO | DKIBO$_{RF}$ | DKIBO$_{GB}$ |
| Colville$_{d=5}$ | 79.59 | **61.66** | 72.79 | 39.52 | **31.61** | 40.32 | **24.52** | 31.00 | 24.90 |
| Michalewicz$_{d=10}$ | 20.1 | 2.33 | **1.63** | 1.9 | **1.53** | 1.67 | 2.15 | **1.97** | 2.05 |
| Ackley$_{d=2}$ | 2.06 | 1.53 | **0.66** | 0.280 | **0.10** | 0.20 | 0.13 | 0.13 | **0.11** |
| Branin$_{d=2}$ | 3.01×10$^{-4}$ | 2.86×10$^{-4}$ | **1.77×10$^{-4}$** | **6.63×10$^{-5}$** | 6.73×10$^{-5}$ | 7.58×10$^{-5}$ | 1.00×10$^{-4}$ | **4.91×10$^{-5}$** | 7.72×10$^{-5}$ |
| Eggholder$_{d=2}$ | 7.66 | 7.54 | **4.87** | 3.04 | **1.11** | 1.23 | 0.48 | 0.27 | 0.03 |
| GoldsteinPrice$_{d=2}$ | 6.02 | 5.64 | **5.24** | 3.04 | **2.85** | 3.58 | **1.14** | 1.64 | 1.84 |
| Hartmann6$_{d=6}$ | 0.51 | 0.24 | **0.075** | **0.0017** | 0.012 | 0.0024 | 0.18 | 0.12 | **8.75×10$^{-4}$** |
| Rosenbrock$_{d=2}$ | 0.37 | 0.34 | **0.16** | 0.11 | 0.10 | **0.072** | 0.10 | 0.067 | **0.055** |
| SixHumpCamel$_{d=2}$ | 0.034 | 0.033 | **0.029** | 0.0012 | 3.10×10$^{-4}$ | **3.02×10$^{-4}$** | 8.03×10$^{-5}$ | 7.20×10$^{-5}$ | **5.84×10$^{-5}$** |
| StyblinskiTang$_{d=2}$ | 0.0018 | 0.0017 | **0.0012** | 7.54×10$^{-4}$ | 7.11×10$^{-4}$ | **6.79×10$^{-4}$** | 0.0010 | 5.73×10$^{-4}$ | **3.93×10$^{-4}$** |
| Robotic swimming | 102.37 | 77.08 | **74.18** | **43.00** | 47.67 | 63.65 | **35.81** | 57.96 | 48.71 |

**Table 6**: Median regret performance on all tasks for various $\kappa$ choices in UCB and predictive model choices. All experiments are conducted in 50 trials of 100 iterations. Due to space limitations standard deviation information is not reported here. Best performances for each $\kappa$ choice appear boldfaced.

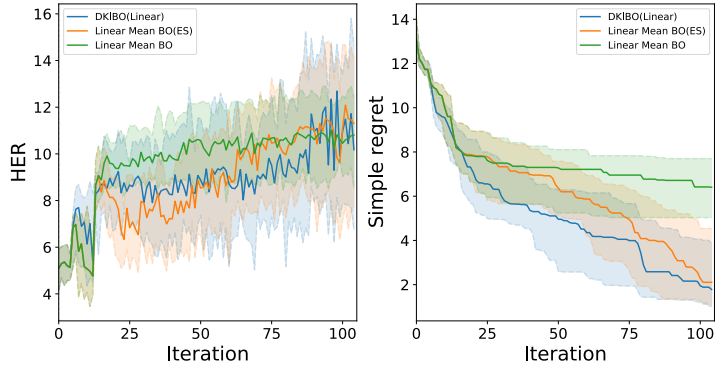

**Figure 4**: HER (left) and simple regret (right) performance comparison on the photocatalysis optimization ablation study between DKIBO and BO using linear mean GP and an early stopping strategy (ES). Solid lines show the median values while shaded areas represent the [25, 75] percentile area.

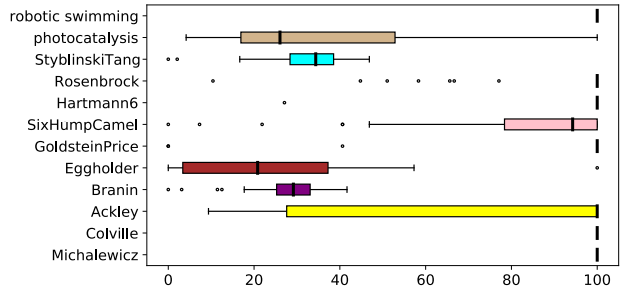

**Figure 5**: Box plots for average usage duration of DKIBO's adaptive corrective term in various problems. Outlier and percentile [25,75] information is also reported.

## 5 Conclusions and future works

In this work we present a novel Bayesian optimization method to inject domain-specific knowledge about the structure of the search space in physical-world problems. We realize this by enriching the acquisition function with a predictive model that adaptively corrects the selection of new sample points in a form of penalty. The proposed approach is simple to implement, generalizable across a wide range of surrogate-based optimization methods and performs competitively on various different settings. Future directions of this work include the development of meta-models that will adaptively control the selection of the predictive model and early stopping criteria in an auto-ML fashion and testing our algorithm on other materials tasks.

## Acknowledgements

The authors acknowledge financial support from the Leverhulme Trust via the Leverhulme Research Centre for Functional Materials Design. Zikai gratefully thanks the China Scholarship Council for a PhD studentship.